\documentclass[sigconf]{acmart}

\AtBeginDocument{%
  \providecommand\BibTeX{{%
    \normalfont B\kern-0.5em{\scshape i\kern-0.25em b}\kern-0.8em\TeX}}}

\usepackage[T1]{fontenc}
\usepackage[utf8]{inputenc}

\renewcommand\footnotetextcopyrightpermission[1]{} 
\usepackage{url}
\usepackage{enumitem}

\usepackage[utf8]{inputenc} 
\usepackage[T1]{fontenc}    
\usepackage{hyperref}       
\usepackage{url}            
\usepackage{booktabs}       
\usepackage{amsfonts}       
\usepackage{nicefrac}       
\usepackage{microtype}      
\usepackage{xcolor}         
\usepackage{subcaption}
\usepackage{graphicx}
\usepackage{svg}
\usepackage{multirow}
\usepackage{wrapfig}
\usepackage{threeparttable}
\usepackage{tikz}

\newcommand*\circled[1]{\raisebox{.4pt}
                    {\tikz[baseline=(char.base)]{
            \node[shape=circle,draw,inner sep=1pt, style={fill=black, text=white}, scale=0.75] (char) {\textbf{#1}};}}}

\begin{document}
\title{Program-to-Circuit: Exploiting GNNs for Program Representation and Circuit Translation}

\author{Nan Wu, Huake He, Yuan Xie}
\affiliation{%
\small
{nanwu,huakehe,yuanxie}@ucsb.edu \\
  \institution{University of California, Santa Barbara}
  \city{Santa Barbara}
  \state{CA}
  \country{USA}
  \postcode{93106}
}

\author{Pan Li}
\affiliation{%
\small
panli@purdue.edu \\
  \institution{Purdue University}
  \city{West Lafayette}
  \state{IN}
  \country{USA}
  \postcode{47907}
}

\author{Cong Hao}
\affiliation{%
\small
callie.hao@ece.gatech.edu \\
  \institution{Georgia Institute of Technology}
  \city{Atlanta}
  \state{GA}
  \country{USA}
  \postcode{30332}
}

\begin{abstract}
Circuit design is complicated and requires extensive domain-specific expertise. 
One major obstacle stuck on the way to \textit{hardware agile development} is the considerably time-consuming process of \textit{accurate circuit quality evaluation}.
To significantly expedite the circuit evaluation during the translation from behavioral languages to circuit designs, we formulate it as a \textbf{Program-to-Circuit} problem, aiming to exploit the representation power of graph neural networks (GNNs) by representing C/C++ programs as graphs.
The goal of this work is four-fold.
\underline{First}, we build a standard benchmark containing 40k C/C++ programs, each of which is translated to a circuit design with actual hardware quality metrics, aiming to facilitate the development of effective GNNs targeting this high-demand circuit design area.
\underline{Second}, 14 state-of-the-art GNN models are tested and analyzed on the Program-to-Circuit problem. We identify key design challenges of this problem, which should be carefully handled but not yet solved by existing GNNs. The goal is to provide domain-specific knowledge for designing GNNs with suitable inductive biases.
\underline{Third}, we discuss three sets of \textit{real-world benchmarks} for GNN generalization evaluation, and analyze the performance gap between the standard programs and the real-case ones.
The goal is to provide more realistic benchmarking, and to enable transfer learning from limited training data to real-world large-scale circuit design problems.
\underline{Fourth}, the Program-to-Circuit problem is a representative within the \textbf{Program-to-X} framework, a set of program-based analysis problems with various downstream tasks. 
The in-depth understanding of strength and weaknesses in applying GNNs on Program-to-Circuit could largely benefit the entire family of Program-to-X.
Pioneering in this direction, we expect more GNN endeavors to revolutionize this high-demand Program-to-Circuit problem and to enrich the expressiveness of GNNs  on programs.
\end{abstract}

\maketitle
\pagestyle{plain}

\section{Introduction}

Graph Neural Networks (GNNs) have achieved significant advancements in representation learning on graph structured data, e.g., particle and high energy physics~\cite{shlomi2020graph, ju2020graph},
chemical analysis~\cite{coley2019graph, gilmer2017neural},
social networks~\cite{fan2019graph, guo2020deep},
and drug-target prediction~\cite{zhao2021identifying, lim2019predicting}.
Most recently, there has been a surge of interest in GNNs approaches for solving electronic design automation (EDA) problems~\cite{li2020customized, zhang2019circuit, wu2021ironman, khailany2020accelerating}, and compiler-related problems by learning program representations~\cite{shi2019learning}.
In these problems, data are naturally presented as graphs. 
For example, circuits are usually described as netlists displaying gates/transistors and their connections, which can be converted into graphs with vertices as gates/transistors and edges as wires; 
the intermediate representations (IRs) of software programs after compilation can also be represented as graphs~\cite{wu2021ironman,shi2019learning}.

\begin{figure}
  \centering
  \vspace{8pt}
  \includegraphics[width=\linewidth]{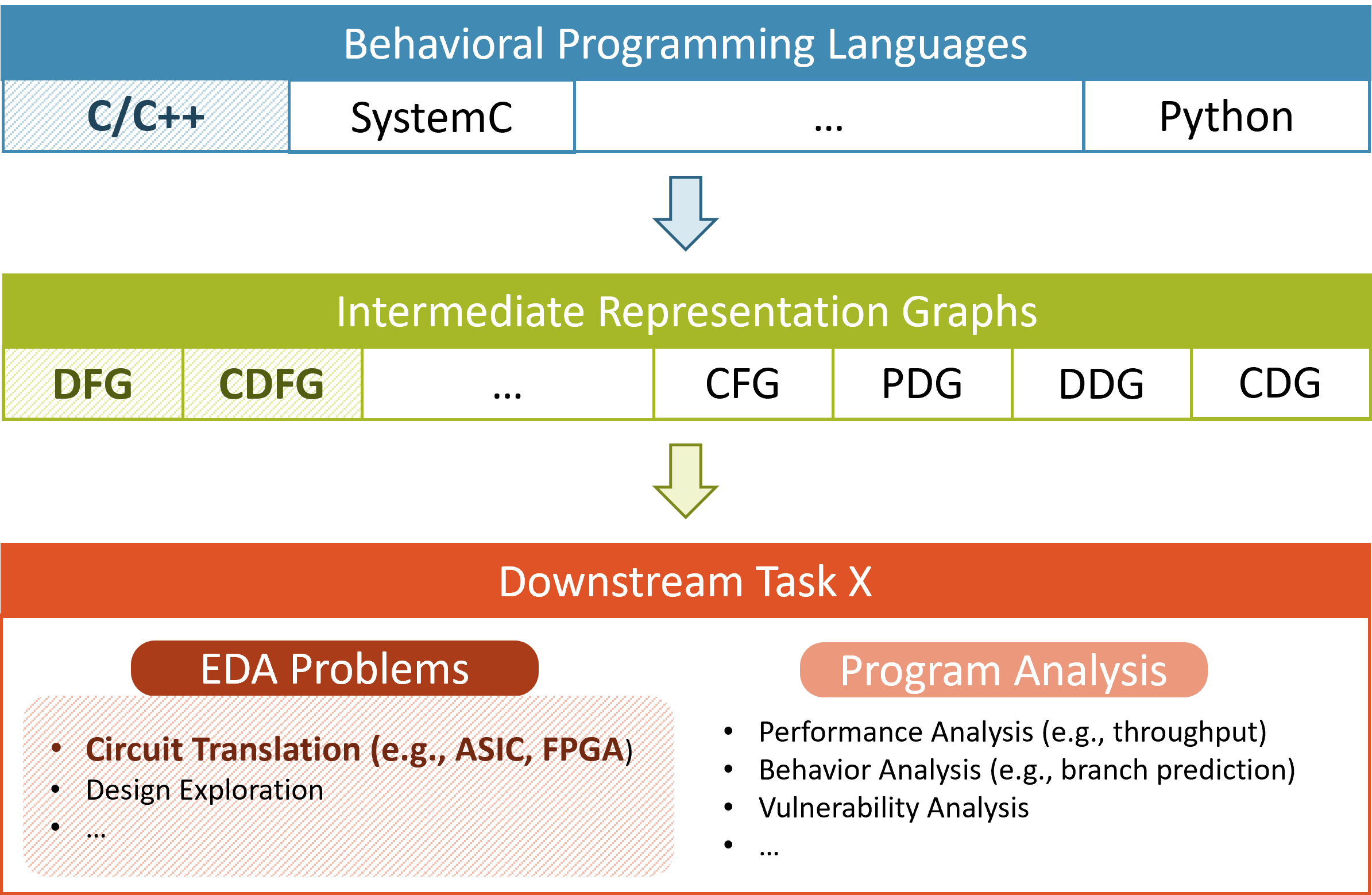}
  \caption{Program-to-Circuit in the scope of Program-to-X.}
  \label{fig:program_to_x}
\end{figure}

Inspired by the expressiveness and representation power of GNNs, we define a new graph representation learning problem in the joint area of EDA and compiler, namely \textbf{Program-to-Circuit}.
Striving for \textit{hardware agile development}, the goal is to significantly expedite circuit quality evaluation and design via GNNs.
More importantly, given a comprehensive study of capabilities and limitations of GNNs on the Program-to-Circuit problem, it would impose greater impacts of GNNs on a much broader domain, the \textbf{Program-to-X} problem.
As depicted in Fig.~\ref{fig:program_to_x}, the Program-to-X framework involves multiple front-ends and downstream tasks, including but not limited to EDA problems and various program analysis.
The Program-to-Circuit problem is one representative in this framework focusing on \textit{fast performance estimation} in the EDA domain, which starts from C/C++ and exploits IR graphs to predict hardware quality after circuit translation.
Other downstream tasks include program analysis on vulnerability \cite{ghaffarian2021neural} or behavior prediction~\cite{shi2019learning}, and CPU throughput estimation~\cite{mendis2019ithemal}, etc.

\textbf{Why Program-to-Circuit matters.} 
Compared with software development, hardware/circuit design is much more time-consuming and requires extensive domain-specific expertise.
Since time to market is a lifeblood for companies to keep competitive, there has been rigid demand for \textit{agile development} of high quality circuits, e.g., application-specific integrated circuits (ASICs) or field-programmable gate array (FPGA) designs, which could eventually reduce the hardware development cycle from years to months or even weeks.
To enable rapid optimization-evaluation iterations in agile development, one essential requirement is to \textit{predict circuit design quality quickly and accurately}.
Traditional EDA tools for circuit design usually take hours to days to accurately predict circuit quality and require extensive manual efforts; 
however, even though high-level synthesis (HLS) tools accelerate the design process by orders of magnitude via automatic translation (a.k.a., synthesis) from behavioral languages into circuit designs, they do require minutes to hours for circuit synthesis and can not accurately estimate circuit quality.
Motivated by the strong desire for hardware agile development and productivity boost, a perfect and sound solution to Program-to-Circuit is expected to bring a huge leap in EDA domain.

\textbf{Why represent programs as graphs.}
First, it is natural and easy for programs to be represented as graphs.
Compilers often use IRs to represent the source code of programs \cite{aho2007compilers, wolf2012computers, gcc, llvm, CPython}, most of which are graph structured, such as \textit{data flow graphs} (DFGs) or \textit{control data flow graphs} (CDFGs). 
Similarly, during the design flow of HLS tools, C/C++ programs are firstly compiled into IR graphs and then translated into circuits.
These IR graphs can be easily and automatically obtained from compiler front-ends (e.g., LLVM~\cite{llvm} and GCC~\cite{gcc}) without manual effort, and they are independent of detailed programming language grammar rules and different standards. 
Second, many performance predictions and design space explorations are operated on DFGs and CDFGs, for example the core HLS algorithms are largely based on the CDFGs~\cite{coussy2009introduction}.

\textbf{From Program-to-Circuit to Program-to-X}.
Given that IR graphs are the common case for representing programs, the in-depth understanding of capabilities and limitations of GNNs on the Program-to-Circuit problem could not only revolutionize this high-demand circuit design area, but also largely benefit the entire family of \textbf{Program-to-X}.
Nonetheless, there are substantial challenges due to limitations in current GNN models. 
First, IR graphs can largely vary in node degree, graph size, and label distribution, and their topologies are further complicated by massive loops, challenging the generalization capability of GNN models, which will be presented in this paper. 
Second, there is a huge gap in representation power required by classification and regression. 
The classification is to predict a discrete class label output for an example, while the regression aims to predict a continuous quantity output (and Program-to-Circuit is an example), which obviously needs more expressiveness in mapping functions. 
Another obstacle is the lack of standard benchmarks. 
Thus, we propose the first benchmark for the Program-to-Circuit problem, and encourage more follow-up benchmark works for various downstream tasks.
We believe that addressing above challenges of GNNs on \textbf{Program-to-Circuit} will light up the way to follow-up researches for \textbf{Program-to-X}.

\textbf{Contributions.}
To address the challenges of agile development for circuit design and to understand the GNN representation power on program IR graphs, in this work, we propose a standard formulation of \textbf{Program-to-Circuit} from \textit{C/C++ programs} by representing them as DFGs and CDFGs, and exploit GNNs for solving the problem.
We summarize our contributions as follows:
\begin{itemize}
    \item {\textbf{Problem Definition.} We are the first to systemically define the Program-to-Circuit problem. It broadens the application of GNNs to an extremely important problem in the hardware design community and challenges the representation power of GNNs through hardware-related features. More importantly, the understanding of GNN expressiveness on general IR graphs could benefit broader program-related problems, i.e., \textbf{Program-to-X}.
    }
    \item {\textbf{Benchmark.} We build a standard benchmark containing 40k C/C++ programs, each of which is translated to a circuit design with actual hardware quality metrics, aiming to facilitate development of effective GNNs targeting this high-demand circuit design area and provide infrastructure to build benchmarks on various program-related downstream tasks.
    }
    \item {\textbf{SOTA GNN analysis.} We test and analyze 14 state-of-the-art GNN models on the Program-to-Circuit mapping. Specifically, We identify key design challenges of this problem, which should be carefully handled but not yet solved by existing GNNs. The goal is to provide domain-specific knowledge for designing GNNs with suitable inductive biases.
    }
    \item {\textbf{Real-case generalization analysis.} We discuss three sets of \textit{real-world benchmarks} for GNN generalization evaluation, and analyze the performance gap between the standard programs and the real-case ones. The goal is to provide pointers for future GNN development along this promising Program-to-X direction.
    }
\end{itemize}

\section{Related Work}

\textbf{GNNs in circuit design.}
There has been signs of emergence in using GNN approaches to solving problems in circuit designs that can be naturally modeled as graphs.
Circuit-GNN~\cite{zhang2019circuit} uses GNNs to simulate electromagnetic (EM) properties of distributed circuits to replace traditional EM simulators, where each node refers to a resonator and each edge refers to the EM coupling between a pair of resonators.
Another category is to model the chip placement into graph formulation and use GNNs to predict the placement quality, where the netlist of the circuit is encoded into a directed graph with cells/components as nodes and the connections between devices as edges.
The prediction goals can be wirelength or chip area for digital circuits~\cite{mirhoseini2020chip}, and gain, bandwidth, or phase margin for analog circuits~\cite{li2020customized}.
Analog circuit (transistor) sizing and symmetry annotation have also been studied~\cite{wang2020gcn, gao2021layout, chen2021universal}, where the constraints for circuit layout are automatically annotated by GNNs.
These prior arts demonstrate potentials of GNNs in circuit design.

\textbf{ML-based program analysis.}
Program (code) analysis has been attracting great research interests for decades~\cite{andersen1994program, nielson2004principles}.
Recently, Xue \textit{et al.}~\cite{xue2019machine} comprehensively survey ML-based approaches for program behavior and performance prediction.
For instance, Ithemal~\cite{mendis2019ithemal} uses a recurrent-neural-network-based architecture with long short term memory (LSTM) for predicting the number of clock cycles that a processor takes to execute a block of assembly instructions (i.e., throughput) on a x86-64 instruction set architecture.
Realizing that IRs of programs can be constructed as graphs, GNNs are employed for different purposes of program analysis.
For example, Shi \textit{et al.~}\cite{shi2019learning} selectively build a graph from the assembly code and perform branch prediction with dynamic states of a program (i.e., the value change of a fixed set of registers) as additional inputs.
Ghaffarian and Shahriari~\cite{ghaffarian2021neural} apply GNNs on different IR graphs, such as control-flow graphs and abstract syntax trees, to detect program vulnerability.
IronMan~\cite{wu2021ironman} is a most recent work where part of it exploits GNNs for both circuit design and program analysis, closest to the Program-to-Circuit problem. It first translates programs into DFGs, and then applies GNNs to predict hardware performance.
Overall, there are rapidly increasing efforts in exploring GNNs for program representation learning and analysis.

\section{Program-to-Circuit Problem}
\label{sec:p2c}

\begin{figure*}
\vspace{8pt}
    \centering
    \includegraphics[width=\textwidth]{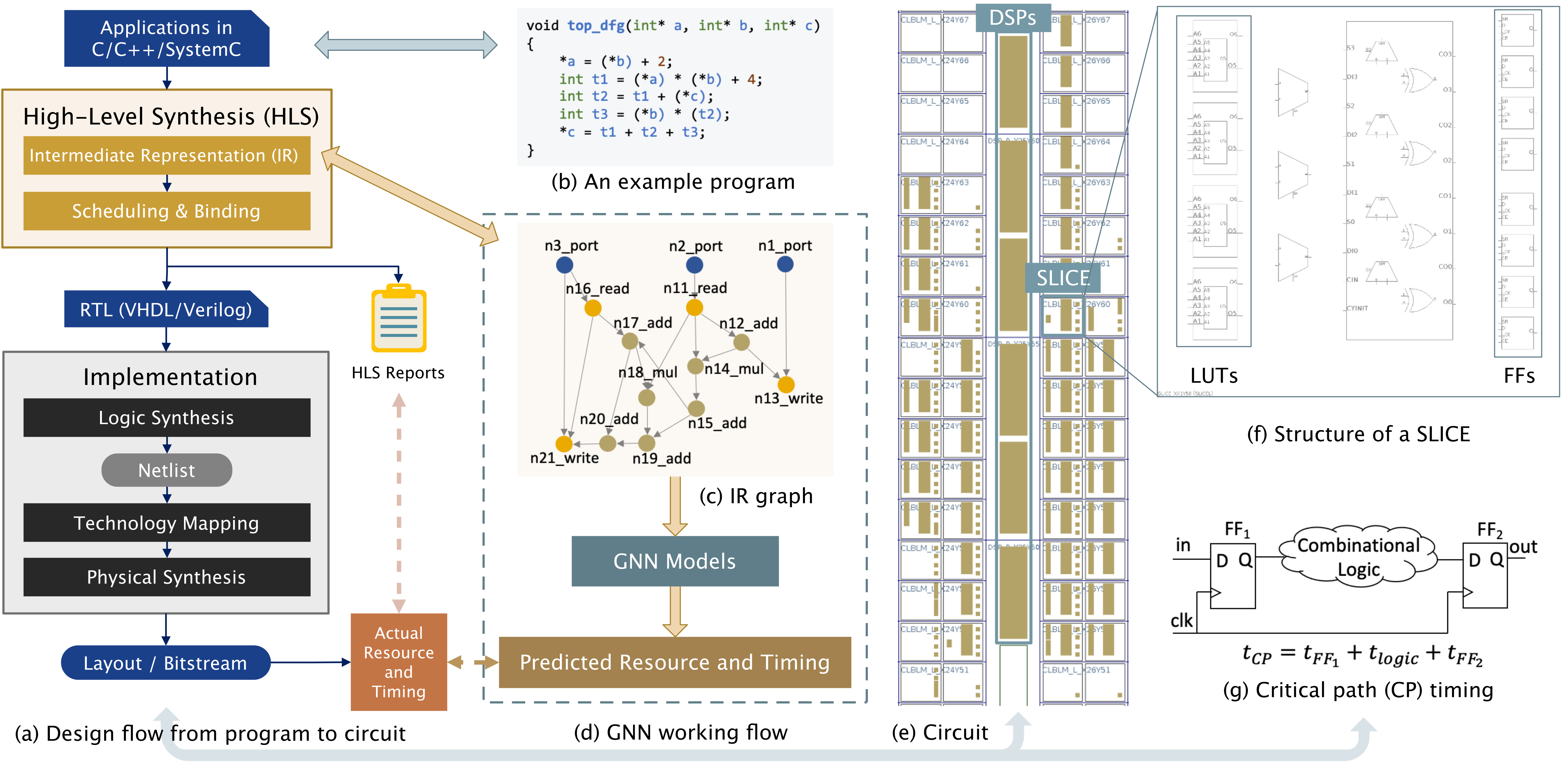}
    \caption{The Program-to-Circuit problem solved in this work. 
    (a) The design flow starting from behavioral programs to hardware circuits. 
    (b) An example program written in C++.
    (c) The intermediate representation (IR) graph extracted by front-end compilers.
    (d) The working flow of GNNs, predicting \textit{actual} resource utilization and timing merely based on raw IR graphs.
    (e) The translated circuit from the example program, using resources including DSPs, LUTs, FFs, and SLICEs. 
    (f) The detailed structure of a SLICE containing LUTs and FFs. 
    (g) An example of critical path (CP) timing.
    HLS tools estimate resource utilization and timing based on RTL designs \textit{after} circuit translation. GNNs are expected to \textit{learn the mapping rules} (including scheduling/binding, logic and physical synthesis), so as to predict the \textit{actual} resource usage and CP timing as accurate as possible.
    }
    \label{fig:Program-to-Circuit}
\end{figure*}

\subsection{Graph-Structured Intermediate Representation}
An intermediate representation (IR) is the data structure used internally by a compiler to represent the source code of a program, which can be used for further processing, such as optimization and translation~\cite{aho2007compilers, wolf2012computers}.
Many modern compilers transform a program into a graph-structured IR, such as the GNU Compiler Collection~\cite{gcc} and LLVM~\cite{llvm} for C/C++, and the CPython~\cite{CPython} interpreter for C and Python.
There are various types of IR graphs, such as data flow graph (DFG), control flow graph (CFG), control data flow graph (CDFG), program dependence graph (PDG), control dependence graph (CDG), and data dependence graph (DDG).
Some graphs can be constructed from others, for example, the CDFG is composed of DFG and CFG, while the PDG is composed of CDG and DDG.

In this work, we consider DFGs and CDFGs as the IR graphs.
Specifically, DFGs are the graphs translated from \textit{basic blocks}, a straight-line code sequence with no branches in except to the entry and no branches out except at the exit~\cite{hennessy2011computer};
CDFGs are the graphs translated from programs with loops \footnote{To distinguish loops in programs and loops in graphs, we use \textit{control loops} in the following context to represent loops in programs.}, jumps, and branches.
The major difference between DFGs and CDFGs is that, DFGs do not contain any loops, while CDFGs contain additional nodes and edges/loops for control dependency.

These IR graphs can be easily obtained by common compiler front-ends, such as LLVM and HLS tools, where HLS tools can be regarded as specialized hardware compilers. 
More realistic examples of programs and their extracted DFGs and CDFGs can be found in the Appendix.

\subsection{Problem Formalization}

\textbf{Input representation.}
The inputs to the Program-to-Circuit problem are directed, possibly cyclic IR graphs representing programs written in behavior-level programming languages, such as C / C++ / SystemC.
These IR graphs are automatically constructed from common compiler front-ends.
Each graph node belongs to one of the three categories: normal operations, blocks (i.e., control signals) or ports (i.e., function arguments).
Graph edges reflect either data dependency or control dependency among nodes.

\textbf{Output.}
With the help of HLS tools, behavioral programs can be directly translated and mapped to circuit designs, either ASICs or FPGA designs.
The functionality of the circuit is exactly the same as the software program, e.g., a fast Fourier transform or a convolutional neural network.
Such transformations follow a set of complex but deterministic rules according to the transformation tools, e.g., program compilers and HLS tools.
Although these tools differ in transformation details across different platforms, they share similar heuristics and mapping rules in both scheduling/binding and logic/physical synthesis.
\textbf{The goal of this work is to exploit GNNs to \textit{learn such underlying heuristics and mapping rules}} given the IR graphs that represents behavior-level programs,
so as to quickly predict the circuit quality (e.g., chip area, resource usage, timing, total power, etc.) as accurately as possible.

\textbf{Case study problem.}
As a case study, we take FPGAs as the target circuit platform.
The translation and mapping flow of C/C++ program to circuits on FPGA is illustrated in Fig.~\ref{fig:Program-to-Circuit}.
GNNs are expected to predict common hardware quality metrics, include resource usage and circuit timing (which determines its maximum working frequency).
We consider five prediction tasks: \texttt{DSP}, \texttt{FF}, \texttt{LUT}, \texttt{SLICE}, \texttt{CP}, as summarized in Table \ref{table:resource}.
The first four are different types of resource, and the last one is circuit timing.
Fig.~\ref{fig:Program-to-Circuit} (e)-(g) provide structures of various resources and CP timing. 

\begin{table*}[tb]
\caption{Prediction tasks of resource and timing on FPGAs.}
\label{table:resource}
\centering
    \renewcommand{\arraystretch}{1}
    \setlength{\tabcolsep}{3pt}
\begin{tabular}{c | c}
\toprule
\textbf{Resource and Timing} & \textbf{Description}  \\ \midrule
\texttt{DSP}: Digital Signal Processor  &  A small processor able to quickly perform mathematical operation on streaming digital signals. \\ 
\texttt{FF}: Flip-Flop                  &  A small memory component able to store a bit, typically used as a fast register to store data. \\ 
\texttt{LUT}: Look-Up Table             &  A set of logic gates hard-wired on FPGAs, storing predefined truth tables and performing logic functions.  \\
\texttt{SLICE}                          &  An elementary programmable logic block that contains a set number of LUTs, FFs and multiplexers.  \\ 
\texttt{CP}: Critical Path Timing       &  The maximum signal delay of a path from an input to an output, usually in the unit of nanoseconds.   \\
\bottomrule
\end{tabular}
\end{table*}

\textbf{Program-to-X extension.}
Keeping the front-end graph-structured inputs unchanged while replacing the downstream prediction tasks can easily extend to other program-related problems within this framework.
For instance, the software vulnerability detection problem~\cite{ghaffarian2021neural} adopts similar input graphs, where the downstream task is a binary classification to decide whether the program is vulnerable or secure.

\subsection{Benchmark}
\textbf{Benchmark generation.}
The benchmark for this Program-to-Circuit problem consists of synthetic C/C++ programs and real-world HLS applications.
The synthetic programs fall into two categories, basic blocks that derive DFGs, and programs with control loops and branches that derive CDFGs.
All of the synthetic programs are generated by a C program generator \texttt{ldrgen} \cite{barany2017liveness}, a plugin of Frama-C \cite{cuoq2012frama,kirchner2015frama}.
There are 19,120 and 18,570 C programs in the DFG dataset and CDFG dataset, respectively.
Each sample is arranged in the form of < program, extracted IR graph, actual resource/timing >.

In addition, we include three sets of real-world HLS applications: MachSuite~\cite{reagen2014machsuite}, CHStone~\cite{hara2009proposal}, and PolyBench/C~\cite{PolyBench}, consisted of 16, 10, and 30 different applications, respectively.
The real-world applications are used for generalization evaluation of different GNN models.

\textbf{Node/Edge Features}.
There are seven features available for each node, which capture both node properties and the paths that this node participates in.
Table \ref{table1} summarizes all the node features with example values.
Each edge has two features, the discrete edge type in integers, and a binary signal marking whether this edge is a back edge.

By reusing the front-end IR graph construction in this work, we expect more follow-up benchmarks with different downstream tasks to be integrated into the Program-to-X framework.

\begin{table*}[tp]
\caption{Node features and example values.}
\label{table1}
\centering
    \renewcommand{\arraystretch}{1}
    \setlength{\tabcolsep}{16pt}
\begin{tabular}{c | c | c }
\toprule
\textbf{Feature} & \textbf{Description} & \textbf{Values} \\ \midrule
Node category     &  General node type & \texttt{operation nodes}, \texttt{blocks}, \texttt{ports}, \texttt{misc}\\ 
Bitwidth          &  Bitwidth of the node  & \texttt{0}$\sim$\texttt{256}, \texttt{misc}\\
Opcode category   &  Opcode categories based on LLVM  & \texttt{binary\_unary}, \texttt{bitwise}, \texttt{memroy}, etc.  \\ 
Opcode            &  Opcode of the node &  \texttt{load}, \texttt{add}, \texttt{mux}, \texttt{xor}, \texttt{icmp}, \texttt{select}, etc.  \\
Is start of path  &  Whether the node is the starting node of a path & \texttt{0}, \texttt{1}, \texttt{misc}   \\
Is LCD node       &  Whether the node uses LCD resource  & \texttt{0}, \texttt{1}, \texttt{misc} \\
Cluster group number  & Cluster number of the node & \texttt{-1}$\sim$\texttt{256}, \texttt{misc}\\
\bottomrule
\end{tabular}
\end{table*}

\subsection{Statistics of Benchmark}

\begin{figure}[tbp]
    \centering
    \includegraphics[width=\linewidth]{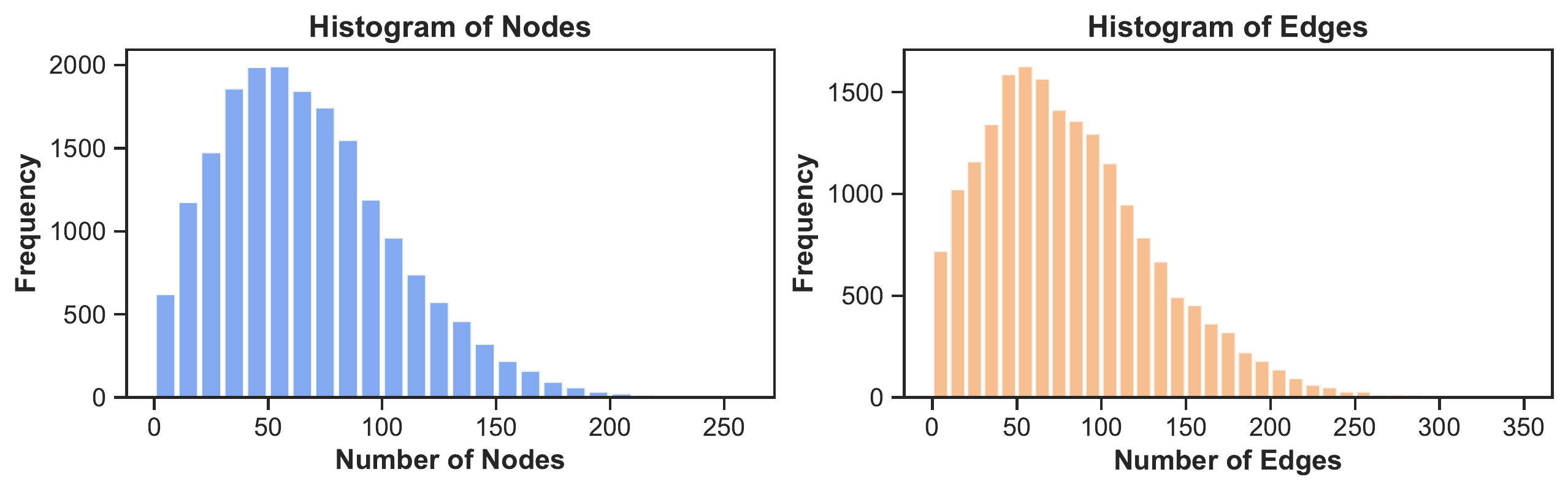}
    \caption{Statistics of the number of nodes/edges in the DFG dataset.}
    \label{fig:dfg_dataset}
\end{figure}

\begin{figure*}[!htbp]
    \centering
    \includegraphics[width=\linewidth]{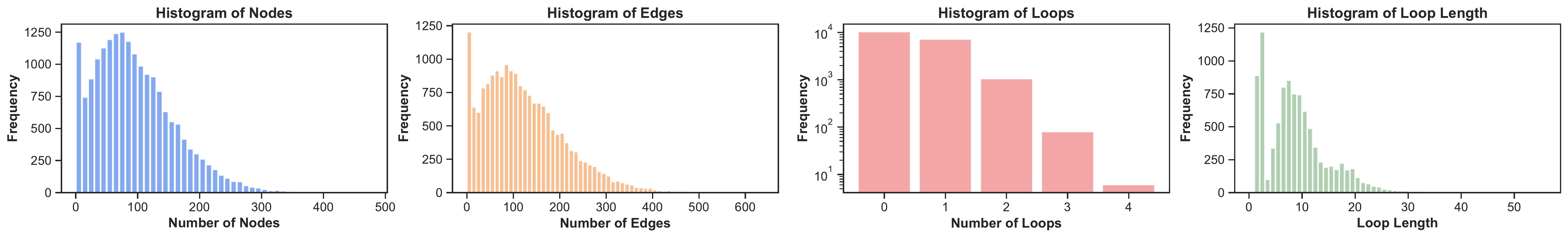}
    \caption{Statistics of the number of nodes/edges and loop information in the CDFG dataset.}
    \label{fig:cdfg_dataset}
\end{figure*}

\textbf{Statistics of Synthetic DFGs/CDFGs.}
Figure \ref{fig:dfg_dataset} and Figure \ref{fig:cdfg_dataset} show the distribution of the number of nodes/edges among 19,120 DFGs in the DFG dataset and among 18,570 CDFGs in the CDFG dataset, respectively.
\begin{itemize}
    \item The graph size of CDFGs is roughly two times as large as that of DFGs.
    \item There is no loop in DFGs since they are generated from basic blocks, while the majority of CDFGs has loops (due to control signals in programs) and some of the loops possess considerable loop length.
\end{itemize}

\begin{figure*}[!htbp]
    \centering
    \includegraphics[width=\linewidth]{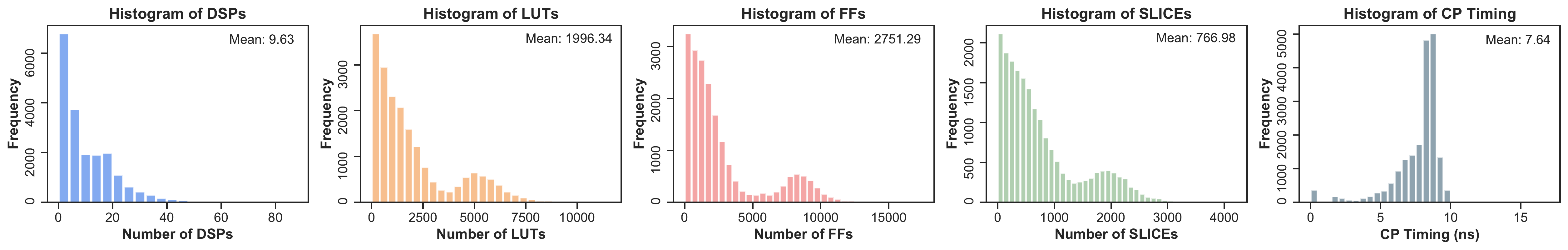}
    \caption{Statistics of resource utilization and timing in the DFG dataset.}
    \label{fig:dfg_resource}
\end{figure*}

\begin{figure*}[!htbp]
    \centering
    \includegraphics[width=\linewidth]{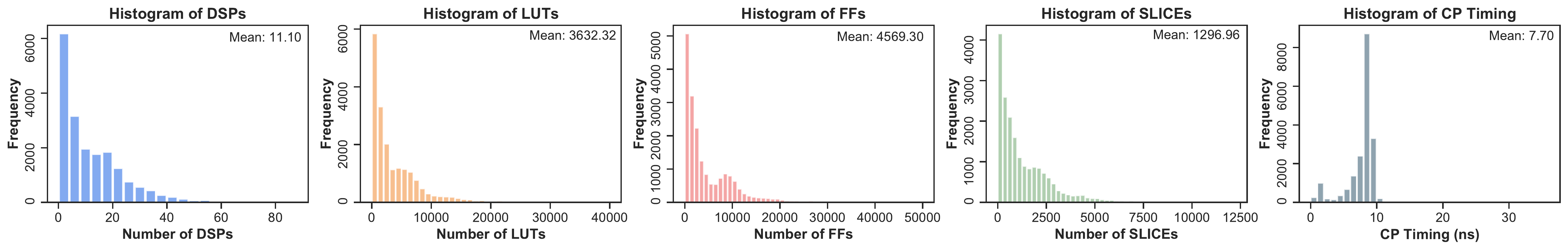}
    \caption{Statistics of resource utilization and timing in the CDFG dataset.}
    \label{fig:cdfg_resource}
\end{figure*}

\textbf{Statistics of Label Values in Synthetic DFGs/CDFGs}
Figure \ref{fig:dfg_resource} and Figure \ref{fig:cdfg_resource} show the distribution of resource usage and timing among 19,120 DFGs in the DFG dataset and among 18,570 CDFGs in the CDFG dataset, respectively.
\begin{itemize}
    \item The CP timing is relatively insensitive to graph sizes, compared with resource utilization.
    \item Among four types of resource utilization, DSP scales with graph sizes, but not significantly.
    \item LUT, FF, and SLICE scale with graph sizes, and they perform a much wider range.
\end{itemize}
    
\begin{figure*}[!htbp]
    \centering
    \includegraphics[width=\linewidth]{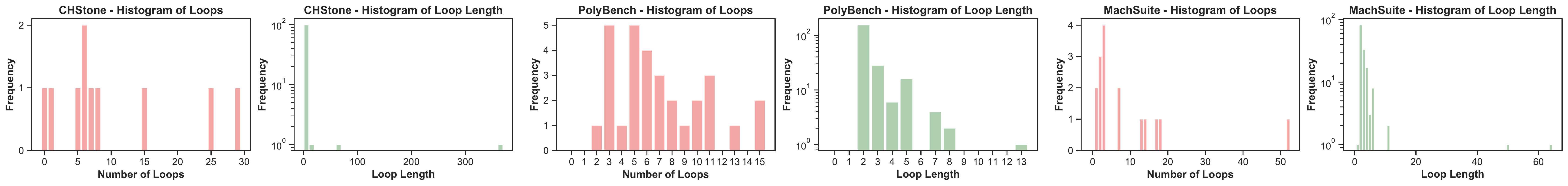}
    \caption{Statistics of loop information in real-case benchmarks.}
    \label{fig:realcase_loop}
\end{figure*}

\textbf{Statistics of Real-world Benchmarks}
Figure \ref{fig:realcase_loop} shows the loop information in real-world benchmarks. Notably, real-world applications tend to have more number of loops in graphs, and some loops have extremely long loop length.
    
Table \ref{table:machsuite} shows graph information, resource utilization and timing of MachSuite. The graphs in MachSuite usually have larger size, with more number of nodes and edges.

Table \ref{table:chstone} shows graph information, resource utilization and timing of CHStone. One characteristic of applications in CHStone is that, each application is consisted of multiple functions, whereas our synthetic programs only have one function per program.

Table \ref{table:polybench} shows graph information, resource utilization and timing of PolyBench. Each application in PolyBench has a single function with a proper graph size, which explains the reason why GNN models generalize better on this benchmark compared to other real-world benchmarks.

We have several observations on the real-world benchmarks:
    \begin{itemize}
        \item In real-world applications, there is no significant relationship between graph size and resource usage, such as \texttt{md\_kernel} in Machsuite, \texttt{adpcm\_main} and \texttt{sha\_stream} in CHStone.
        \item Several applications have really complicated operations, such as \texttt{back\_prop} in MachSuite.
    \end{itemize}

\begin{table}[tbp]
\caption{Detailed information of MachSuite.}
\label{table:machsuite}
\centering
    \renewcommand{\arraystretch}{1}
    \setlength{\tabcolsep}{2pt}
\footnotesize
\begin{tabular}{c|ccccccc}
\toprule
\textbf{MachSuite} & \textbf{\# nodes} & \textbf{\# edges} & \textbf{DSP} & \textbf{LUT} & \textbf{FF} & \textbf{SLICE} & \textbf{CP (ns)} \\ \midrule
gemm\_ncubed         & 801  & 1068 & 28  & 5891  & 11427 & 2550  & 8.563  \\
stencil              & 98   & 139  & 27  & 472   & 697   & 248   & 6.317  \\
fft                  & 88   & 144  & 50  & 2174  & 3185  & 998   & 8.822  \\
md\_grid             & 423  & 645  & 134 & 9201  & 14103 & 3723  & 9.754  \\
stencil3d            & 195  & 325  & 6   & 619   & 633   & 228   & 6.889  \\
ellpack              & 155  & 208  & 28  & 2456  & 4280  & 1268  & 8.612  \\
viterbi              & 878  & 3336 & 6   & 21079 & 15910 & 6730  & 10.053 \\
needwun              & 225  & 337  & 0   & 816   & 598   & 273   & 9.023  \\
kmp                  & 93   & 147  & 0   & 243   & 336   & 125   & 6.119  \\
ms\_mergesort        & 79   & 180  & 0   & 693   & 688   & 257   & 7.138  \\
ss\_sort             & 453  & 734  & 0   & 2089  & 2232  & 772   & 7.623  \\
bbgemm               & 184  & 281  & 14  & 1481  & 2215  & 658   & 8.511  \\
md\_kernel           & 598  & 888  & 268 & 36289 & 57959 & 16207 & 9.313  \\
backprop             & 3506 & 5037 & 107 & 46072 & 58091 & 13155 & 9.896  \\
spmv                 & 48   & 64   & 14  & 979   & 1307  & 475   & 7.891  \\
aes256\_encrypt\_ecb & 462  & 1182 & 0   & 1388  & 1143  & 491   & 6.99   \\
\midrule
\textbf{Mean}      & 517.8   & 919.7  & 42.6  & 8246.4     & 10925.25    & 3009.9       & 8.2196  \\
\bottomrule
\end{tabular}
\end{table}

\begin{table}[tbp]
\caption{Detailed information of CHStone.}
\label{table:chstone}
\centering
    \renewcommand{\arraystretch}{1}
    \setlength{\tabcolsep}{2pt}
\footnotesize
\begin{tabular}{c|ccccccc}
\toprule
\textbf{CHStone}    & \textbf{\# nodes} & \textbf{\# edges} & \textbf{DSP} & \textbf{LUT}    & \textbf{FF}     & \textbf{SLICE} & \textbf{CP (ns}) \\ \midrule
aes\_main          & 2173     & 2814     & 0   & 545    & 889    & 250   & 5.549   \\
adpcm\_main        & 1702     & 2335     & 0   & 78     & 165    & 44    & 4.027   \\
sha\_stream        & 1899     & 3002     & 0   & 134    & 365    & 76    & 5.117   \\
float64\_mul       & 337      & 524      & 16  & 2146   & 1249   & 685   & 8.747   \\
Gsm\_LPC\_Analysis & 461      & 789      & 5   & 2430   & 2396   & 847   & 8.337   \\
blowfish\_main     & 521      & 1219     & 0   & 2143   & 3273   & 915   & 8.124   \\
local\_sin         & 1443     & 2359     & 43  & 10294  & 7488   & 3407  & 9.184   \\
mips               & 456      & 890      & 8   & 1300   & 806    & 436   & 7.719   \\
float64\_add       & 325      & 775      & 0   & 4236   & 2121   & 1284  & 8.471   \\
float64\_div       & 421      & 709      & 24  & 3805   & 3529   & 1297  & 9.04    \\ \midrule
\textbf{Mean}               & 973.8    & 1541.6   & 9.6 & 2711.1 & 2228.1 & 924.1 & 7.4315  \\ \bottomrule
\end{tabular}
\end{table}

\begin{table}[tbp]
\caption{Detailed information of PolyBench.}
\label{table:polybench}
\centering
    \renewcommand{\arraystretch}{1}
    \setlength{\tabcolsep}{2pt}
\footnotesize
\begin{tabular}{c|ccccccc}
\toprule
\textbf{PolyBench} & \textbf{\# nodes} & \textbf{\# edges} & \textbf{DSP} & \textbf{LUT} & \textbf{FF} & \textbf{SLICE} & \textbf{CP (ns)} \\ \midrule
kernel\_gramschmidt       & 151 & 221 & 14 & 6550  & 11343 & 2559 & 8.254 \\
kernel\_jacobi\_1d\_imper & 70  & 95  & 14 & 1056  & 1372  & 440  & 8.471 \\
kernel\_doitgen           & 95  & 137 & 21 & 1282  & 1392  & 574  & 8.649 \\
kernel\_correlation       & 237 & 361 & 21 & 7522  & 12618 & 2770 & 8.476 \\
kernel\_floyd\_warshall   & 98  & 168 & 15 & 1523  & 2080  & 636  & 9.177 \\
kernel\_trisolv           & 75  & 119 & 14 & 4393  & 7665  & 1690 & 8.194 \\
kernel\_jacobi\_2d\_imper & 117 & 164 & 21 & 1860  & 2460  & 801  & 8.452 \\
kernel\_cholesky          & 108 & 168 & 14 & 6468  & 11092 & 2491 & 8.697 \\
kernel\_atax              & 93  & 131 & 14 & 1078  & 1464  & 531  & 7.692 \\
kernel\_2mm               & 149 & 214 & 22 & 1698  & 2219  & 753  & 8.246 \\
kernel\_symm              & 102 & 147 & 29 & 1625  & 2489  & 727  & 8.413 \\
kernel\_bicg              & 80  & 110 & 14 & 1092  & 1612  & 512  & 7.825 \\
kernel\_dynprog           & 157 & 239 & 7  & 785   & 1135  & 368  & 6.527 \\
kernel\_covariance        & 155 & 239 & 20 & 4965  & 8380  & 1902 & 8.292 \\
kernel\_ludcmp            & 269 & 403 & 14 & 5763  & 8936  & 2056 & 9.039 \\
kernel\_reg\_detect       & 243 & 363 & 0  & 795   & 1057  & 364  & 7.693 \\
kernel\_seidel\_2d        & 117 & 168 & 6  & 4688  & 5180  & 1539 & 8.781 \\
kernel\_fdtd\_apml        & 377 & 551 & 77 & 18806 & 28764 & 6136 & 9.466 \\
kernel\_durbin            & 153 & 217 & 17 & 1918  & 2183  & 737  & 8.15  \\
kernel\_gemm              & 80  & 110 & 18 & 1240  & 1707  & 556  & 7.87  \\
kernel\_trmm              & 76  & 122 & 18 & 1358  & 1810  & 555  & 8.528 \\
kernel\_syrk              & 106 & 156 & 18 & 1269  & 1871  & 639  & 7.897 \\
kernel\_gemver            & 175 & 253 & 33 & 2613  & 3495  & 1121 & 8.416 \\
kernel\_fdtd\_2d          & 189 & 278 & 40 & 3897  & 4701  & 1618 & 8.681 \\
kernel\_3mm               & 196 & 287 & 26 & 1959  & 2656  & 988  & 7.704 \\
kernel\_gesummv           & 66  & 91  & 37 & 2731  & 2507  & 1134 & 8.625 \\
kernel\_mvt               & 97  & 140 & 14 & 1236  & 1565  & 537  & 7.943 \\
kernel\_lu                & 108 & 180 & 22 & 4867  & 8407  & 1855 & 8.288 \\
kernel\_syr2k             & 116 & 171 & 18 & 1440  & 2119  & 705  & 8.189 \\
kernel\_adi               & 339 & 550 & 22 & 7624  & 10810 & 2769 & 9.236 \\
\midrule
\textbf{Mean}      & 146.5        & 218.4        & 20.7   & 3470.0   & 5169.6  & 1335.4     & 8.3290       \\ 
\bottomrule
\end{tabular}
\vspace{-8pt}
\end{table}

\section{GNN Models}

GNNs operate by propagating information along the edges of a given graph.
Specifically, each node $v$ is initialized with an representation $h_v^{(0)}$, which could be either a direct representation or a learnable embedding obtained from features of this node.
Then, a GNN layer updates each node representation by integrating the node representations of its neighbors in the graph, yielding representations $h_v^{(1)}$.
This process can be unrolled through time steps by repeatedly using the same update function, deriving representations $h_v^{(2)}, h_v^{(3)}, ..., h_v^{(T)}$.
An alternative is to stack several GNN layers, intuitively similar to unrolling through time steps, but increases the GNN capacity by using different parameters in the update function for each time step.

\begin{figure*}[tb]
    \centering
    \includegraphics[width=\textwidth]{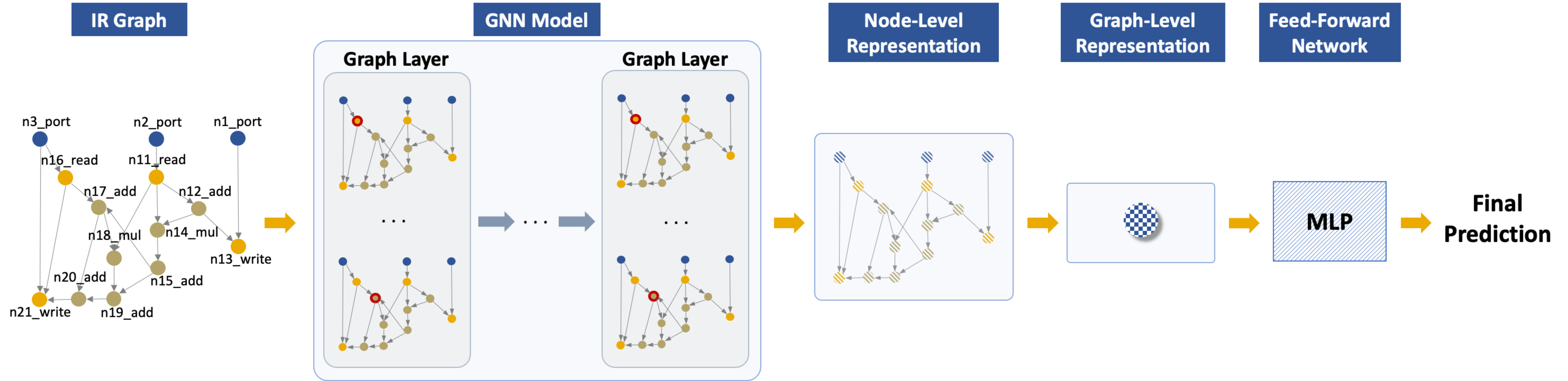}
    \caption{Layout of the structure used. When comparing different models, the difference lies only in the type of graph layers.}
    \label{fig:gnn}
\end{figure*}

In this work, we test 14 different GNN models and analyze their performance on the Program-to-Circuit problem.
To fairly evaluate these models, we use a general structure as shown in Fig. \ref{fig:gnn}.
These 14 GNN models can be roughly categorized into four groups, according to their origins, the way to update node representations, and whether edge information is considered.

\textbf{Graph Convolutional Network (GCN) and variants.}
\circled{1} GCN \cite{kipf2016semi} is inspired by the first order graph Laplacian methods, and it essentially performs aggregation and transformation on node representations without learning trainable filters.
\circled{2} GCN can be equipped with a virtual node \cite{gilmer2017neural}; this virtual node serves as a global scratch space that each node reads from and writes to in every step of message passing.
\circled{3} SGC \cite{wu2019simplifying} is a simplified version of GCN, which reduces computation complexity through successively removing nonlinearities and collapsing weight matrices between consecutive layers, corresponding to a fixed low-pass filter followed by a linear classifier.
\circled{4} GraphSage \cite{hamilton2017inductive} can be recognized as a variant of GCN, which samples a fixed number of neighboring nodes to keep the computational footprint consistent.
\circled{5} The convolution operation based on auto-regressive moving average filters (ARMA) \cite{bianchi2021graph} is able to offer a larger variety of frequency responses and can account for higher-order neighborhoods compared to polynomial filters with the same number of parameters. 
\circled{6} PAN \cite{ma2020path} considers path integral information in the convolution operation, which is a generalization of GCN that assigns trainable weights to each path depending on its length.

\textbf{Graph Isomorphism Network (GIN) and variants.}
\circled{1} GIN \cite{xu2018powerful} is provably as powerful as Weisfeiler-Lehman graph isomorphism test, taking advantage of sum aggregators over a countable input feature space.
\circled{2} Similarly, GIN can also be equipped with a virtual node \cite{gilmer2017neural}.
\circled{3} Principle neighborhood aggregation (PNA) \cite{corso2020principal} emphasizes the necessity to use complementary aggregators, which allows each node to better understand the graph structure and retain neighborhood information, especially under a continuous input feature space. The sum aggregator is generalized as a combination of a mean aggregation and degree-scalers, enabling the network to amplify or attenuate signals based on the degree of each node.

\textbf{Employment of multi-relational information.}
\circled{1} Graph attention networks (GATs) \cite{velivckovic2017graph} apply attention mechanisms to implicitly assign different importances to nodes in the same neighborhood.
\circled{2} Gated graph neural networks (GGNNs) \cite{li2015gated} have trainable edge-dependent weights with gated recurrent units.
\circled{3} Rather than unrolling layer-wise computation through time steps as GAT and GGNN, relational GCN (RGCN) \cite{schlichtkrull2018modeling} utilizes edge-dependent weight with non-linearity activation, which is specifically designed to characterize multi-relational data and contextual information.

\textbf{Inspired from vision tasks.}
\circled{1} Inspired by the advances in pixel-wise prediction tasks brought by encoder-decoder architectures such as the U-Net, graph U-Net \cite{gao2019graph} develops an encoder-decoder structure on graph, which can encode and decode high-level features while maintaining local spatial information.
\circled{2} Inspired by the feature-wise linear modulation (FiLM) in the visual question answering domain, GNN-FiLM \cite{brockschmidt2020gnn} makes use of hypernetworks in learning on graphs, combining learned message passing functions with dynamically computed element-wise affine transformations.

\section{Experiment}

\subsection{Experimental Setup}
\label{sec:exp_setup}
Our framework is built upon Open Graph Benchmark (OGB) library \cite{hu2020ogb}, and the proposed benchmark datasets are compatible with OGB.
All the aforementioned GNN models are implemented with Pytorch Geometric \cite{Fey/Lenssen/2019}.
Experiments were performed on a Linux host with a 64-core Intel Xeon Gold 5218 CPU (2.30 GHz) and Nvidia RTX 2080Ti GPUs.
For DFG and CDFG datasets, the data are randomly split into 80\% train, 10\% validation and 10\% test; real-world benchmarks are only used for generalization evaluation.
Each GNN model is empirically set to have 5 layers, with a hidden-dimension size of 300.
For synthetic dataset, the sum pooling is used to derive graph-level representations.
Since the graph scale of real-world applications are significantly larger than that of synthetic IR graphs, the mean pooling is used to guarantee better generalization.
The feed-forward network has the structure 300-600-300-1.
We train models using the Adam optimizer for 300 epochs.
Learning rates, weight decay, dropout and other hyper-parameters are tuned on the validation set. 
For each model, we conduct five training runs with different random number seeds and hyper-parameters (but close to the tuned values) and report the average of three with least validation error.




\subsection{SOTA GNNs on DFG and CDFG from Synthetic Programs}

\textbf{DFG vs. CDFG}.
Table \ref{table:result} exhibits relative errors of predictions on DFGs and CDFGs from the synthetic programs.
The error rates on CDFGs are conspicuously larger than those on DFGs, due to three major reasons.
First, the graph size of CDFGs is approximately twice as large as that of DFGs, increasing the difficulty of graph-level regression.
Second, DFGs are extracted from basic blocks without controls and branches, so that there are no \textit{loops} in the graph; by contrast, CDFGs has a considerable number of loops, as shown in Fig.~\ref{fig:cdfg_dataset}, challenging the representation power of GNNs.
Third, control signals introduce additional nodes/edges that represent control states and dependency, which are seemingly unrelated to resource usage, greatly encumbering the resource prediction; 
meanwhile, control signals are usually accompanied with more complex memory operations \cite{xilinx}, such as \texttt{store} and \texttt{alloca}, further complicating the allocation of FF/LUT/SLICE.

\begin{figure*}[t]
      \centering
      \begin{subfigure}{\linewidth}
          \centering
          \includegraphics[width=\linewidth]{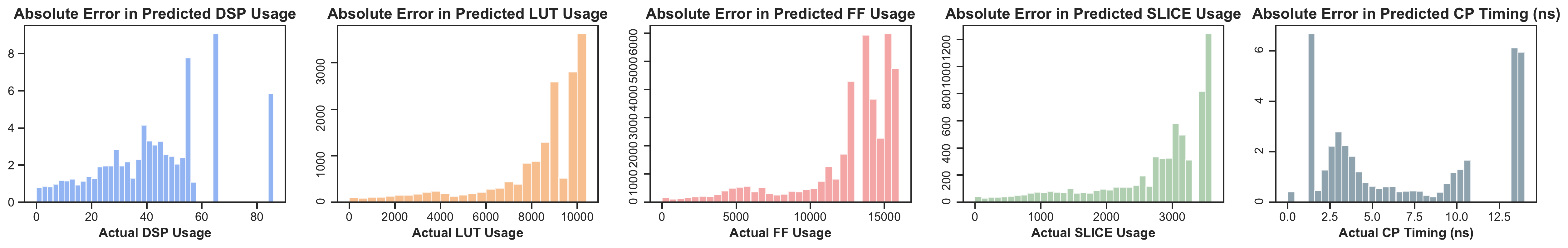}
          \vspace{-15pt}
          \caption{Absolute errors of DSP, LUT, FF, SLICE and CP timing in DFGs.}
          \label{fig:dfg_abs}
      \end{subfigure}
      \begin{subfigure}{\linewidth}
          \centering
          \includegraphics[width=\linewidth]{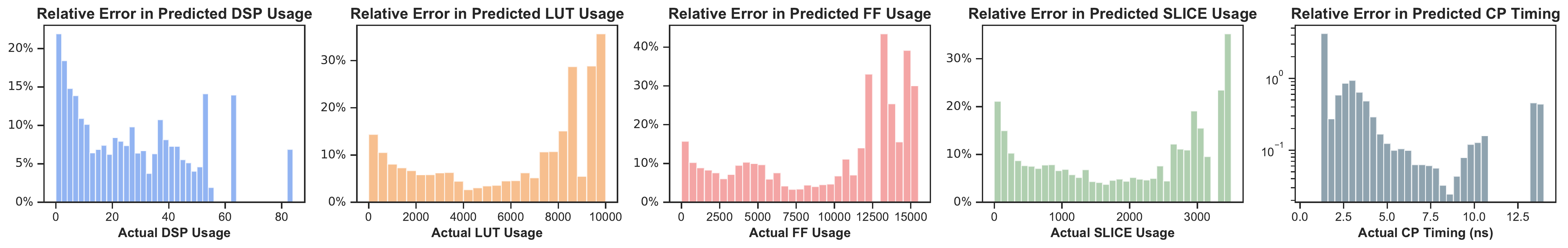}
          \vspace{-15pt}
          \caption{Relative errors of DSP, LUT, FF, SLICE and CP timing in DFGs.}
          \label{fig:dfg_re}
      \end{subfigure}
      \begin{subfigure}{\linewidth}
          \centering
          \includegraphics[width=\linewidth]{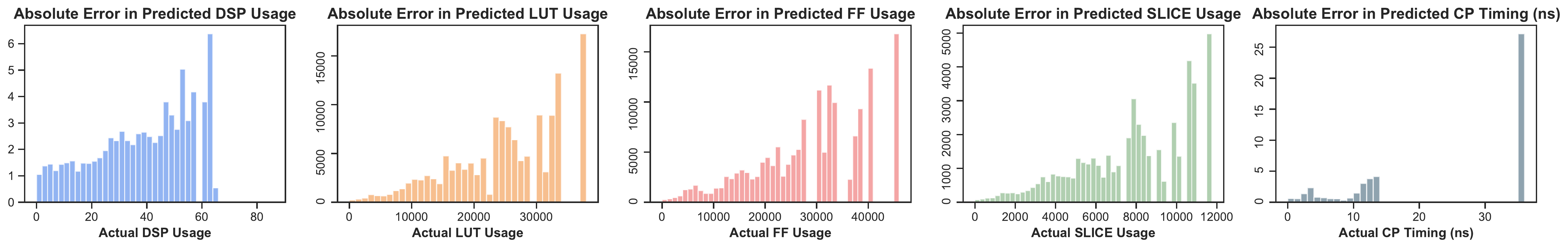}
          \vspace{-15pt}
          \caption{Absolute errors of DSP, LUT, FF, SLICE and CP timing in CDFGs.}
          \label{fig:cdfg_abs}
      \end{subfigure}
      \begin{subfigure}{\linewidth}
          \centering
          \includegraphics[width=\linewidth]{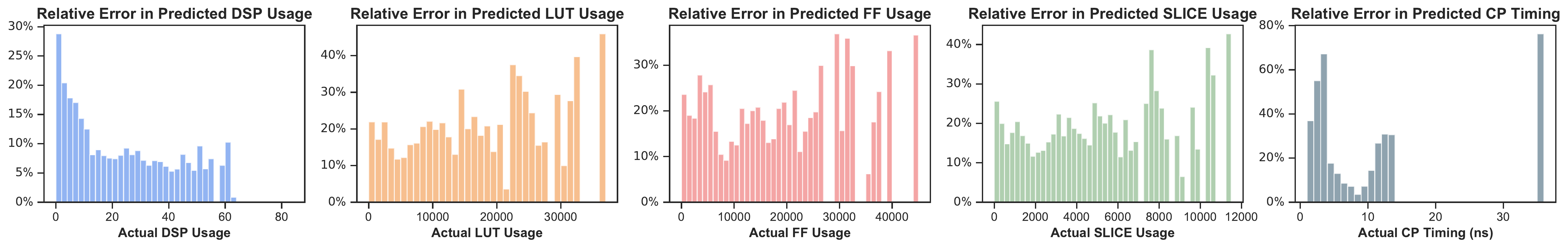}
          \vspace{-15pt}
          \caption{Relative errors of DSP, LUT, FF, SLICE and CP timing in CDFGs.}
          \label{fig:cdfg_re}
      \end{subfigure}
      \caption{Absolute and relative errors of PNA on DFGs and CDFGs.}
      \label{fig:relative}
\end{figure*}

\textbf{Model Analysis}.
In terms of performance variance of different models, PNA and RGCN generally show superior performance, implying two takeaways.
First, the relational information (i.e., edge information) is important in IR graphs, since they represent data dependency, control dependency, or a mix of these two, which is a critical basis in logic synthesis and impacts resource allocation.
Second, equipped with multiple aggregators, PNA is more powerful to characterize different neighborhood information of each node, thus making better predictions.
Fig. \ref{fig:relative} shows how absolute and relative errors of PNA predictions distribute with ground truth.
It is noteworthy that in the relatively small cases (with small resource usage), the absolute errors stay in a rather consistent range, which leads to decrease of relative errors in a multiplicative inverse manner;
in the middle part, the absolute errors grow with the graph sizes, resulting in consistent relative errors;
in the relatively large cases, the incapability of generalization across graph size is exposed.

\begin{table*}[tb]
\caption{Relative errors of predictions on DFG and CDFG datasets, among different GNN models. The top two models with least error rates are marked in bold.}
\label{table:result}
\centering
    \renewcommand{\arraystretch}{0.9}
    \setlength{\tabcolsep}{4pt}
\begin{tabular}{c|ccccc|ccccc}
\toprule
                   & \multicolumn{5}{c|}{DFG}   & \multicolumn{5}{c}{CDFG}       \\ 
                   & DSP      & LUT      & FF       & SLICE    & CP Timing 
                   & DSP      & LUT      & FF       & SLICE    & CP Timing       \\ \midrule
GCN                & 16.31\%  & 16.49\%  & 21.27\%  & 22.29\%  & \textbf{6.12\%} 
                   & 25.30\%  & 28.64\%  & 38.34\%  & 43.63\%  & 8.79\%          \\
GCN - Virtual Node & 15.72\%  & 15.93\%  & 21.64\%  & 23.21\%  & 6.36\%          
                   & 17.31\%  & 33.93\%  & 39.94\%  & 49.22\%  & \textbf{8.13\%} \\
SGC                & 42.12\%  & 23.93\%  & 30.61\%  & 28.20\%  & 7.92\%          
                   & 44.01\%  & 60.87\%  & 53.50\%  & 64.10\%  & 10.32\%         \\
GraphSage          & 15.18\%  & 14.01\%  & 17.11\%  & 15.90\%  & \textbf{6.12\%} 
                   & 17.01\%  & 28.09\%  & 39.11\%  & 46.53\%  & \textbf{8.25\%} \\
ARMA               & 19.12\%  & 13.46\%  & 16.87\%  & 16.09\%  & 6.50\%          
                   & 18.47\%  & \textbf{25.21\%} & 32.15\% & 28.31\% & 8.42\%    \\
PAN                & 15.24\%  & 14.13\%  & 17.23\%  & 16.49\%  & 6.38\%          
                   & 16.88\%  & 32.65\%  & 44.36\%  & 44.84\%  & 8.54\%          \\
GIN                & 15.52\%  & 16.10\%  & 22.08\%  & 22.63\%  & 6.58\%          
                   & 15.47\%  & 28.48\%  & 38.82\%  & 46.12\%  & 8.76\%          \\
GIN - Virtual Node & 15.04\%  & 16.17\%  & 23.09\%  & 24.19\%  & 6.40\%          
                   & 17.94\%  & 29.40\%  & 48.64\%  & 49.44\%  & 8.59\%          \\
PNA  & \textbf{12.65\%} & \textbf{11.64\%} & \textbf{14.41\%} & \textbf{14.34\%} & 6.26\% 
     & \textbf{14.71\%} & \textbf{22.86\%} & \textbf{26.47\%} & \textbf{23.38\%} & 8.87\% \\
GAT                & 26.22\%  & 22.64\%  & 27.74\%  & 25.68\%  & 8.30\%  
                   & 28.66\%  & 46.19\%  & 54.73\%  & 64.72\%  & 10.32\%         \\
GGNN               & 15.40\%  & 13.64\%  & 16.94\%  & 15.79\%  & 6.47\%          
                   & 16.28\%  & 28.05\%  & 31.88\%  & 34.15\%  & 8.50\%          \\
RGCN & \textbf{13.27\%} & 13.03\% & \textbf{15.09\%} & \textbf{15.11\%} & 6.14\% 
     & \textbf{15.03\%} & 26.33\% & \textbf{25.52\%} & \textbf{22.84\%} & 8.72\% \\
UNet               & 18.40\%  & 14.90\%  & 19.17\%  & 17.18\%  & 6.61\% 
                   & 18.92\%  & 32.83\%  & 53.06\%  & 54.35\%  & 9.02\%          \\
FiLM               & 20.05\%  & \textbf{12.50\%} & 16.94\% & 16.30\% & 6.27\% 
                   & 17.42\%  & 26.97\%  & 27.35\%  & 31.50\%  & 8.67\%         \\
\bottomrule
\end{tabular}
\end{table*}

\begin{figure*}[tbp]
    \centering
    \begin{subfigure}{\linewidth}
        \centering
        \vspace{-35pt}
        \includegraphics[width=0.92\linewidth]{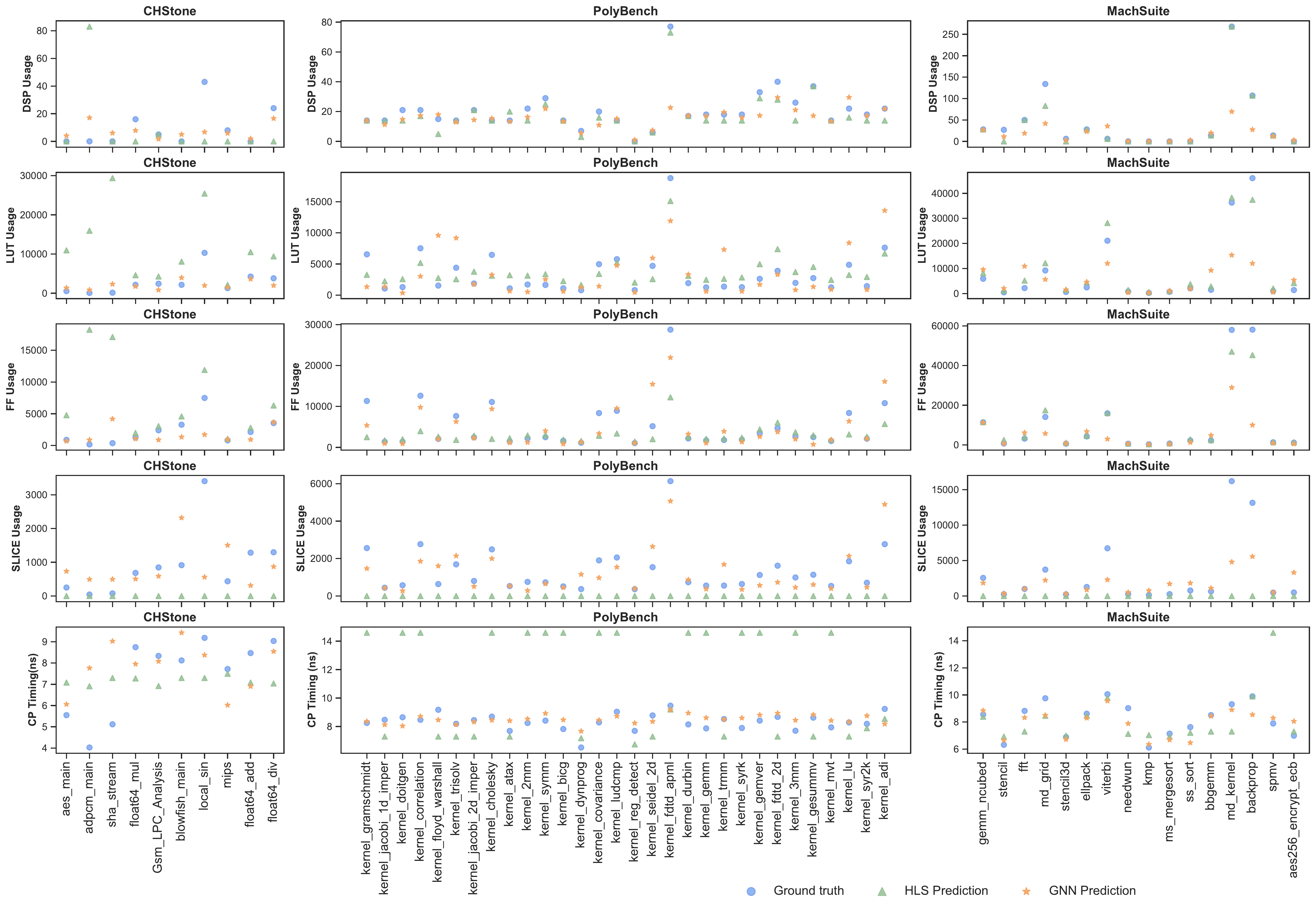}
        \vspace{-5pt}
        \caption{Predictions of PNA on three real-world benchmarks.}
        \label{fig:realcase_abs}
    \end{subfigure}
    \begin{subfigure}{\linewidth}
        \centering
        \includegraphics[width=0.90\linewidth]{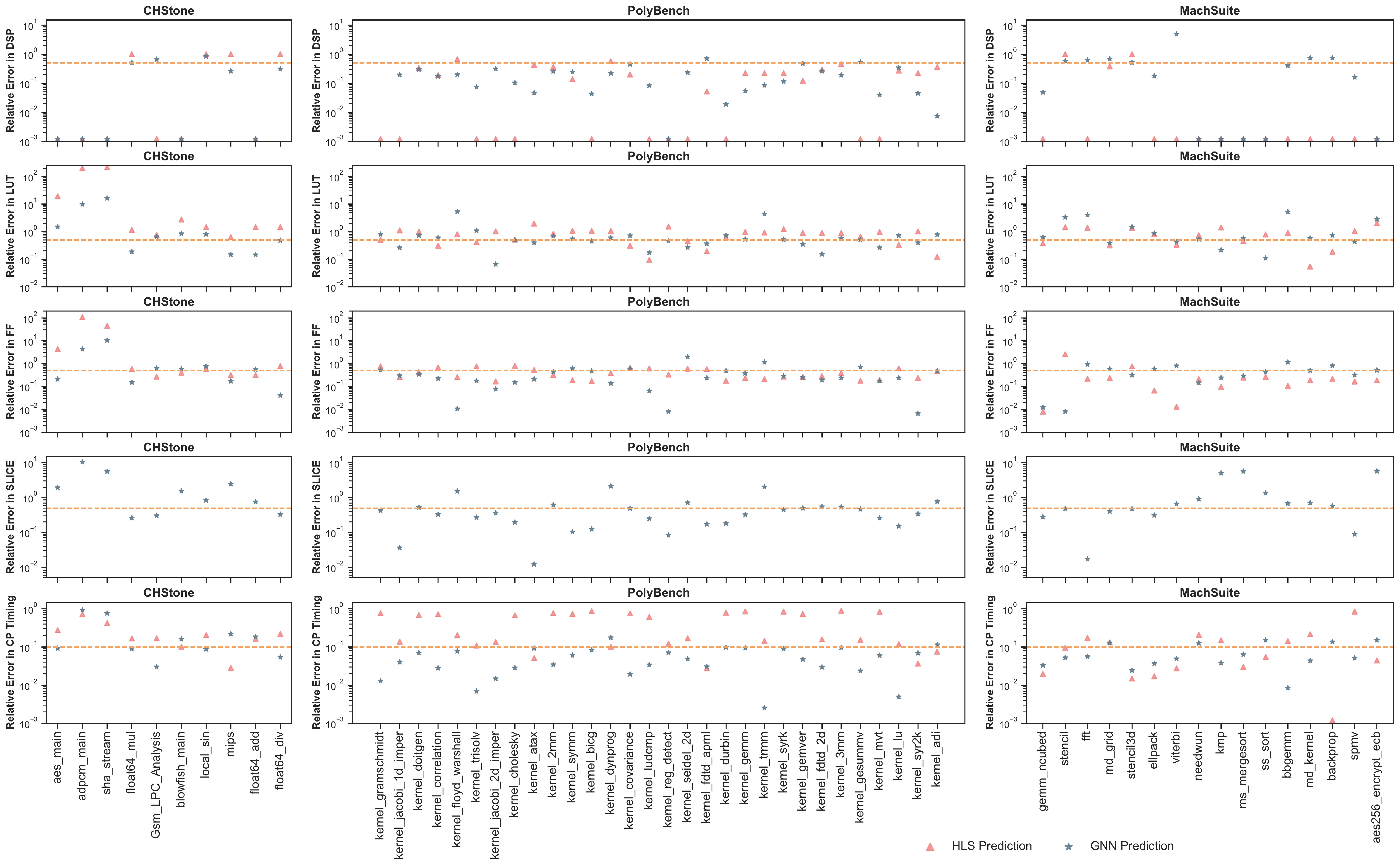}
        \vspace{-5pt}
        \caption{Relative prediction errors of PNA on three real-world benchmarks.}
        \label{fig:realcase_re}
    \end{subfigure}
    \caption{Generalization of PNA on three real-world benchmarks. Note that in Figure \ref{fig:realcase_re}, the stars and triangles that reside on the horizontal axis indicate either the ground truth to be zero or perfect predictions. Since HLS is not able to provide predictions on SLICE utilization, the corresponding relative error is left blank.}
    \label{fig:realcase}
\end{figure*}

\textbf{Domain-specific Insights}.
\circled{1} Resource.
Among four types of resource, DSP is mainly used for computation; 
FF utilization often relates to memory operations and small arrays;
LUTs may appear in computation, memory or control nodes; 
SLICE is a functional unit that comprises multiple LUTs and FFs.
The core to making precise DSP prediction is to distinguish major computation nodes that are most likely to use DSPs. 
For instance, a multiplication node with a large bitwidth tends to use DSPs, while divisions and bitwise operations prefer LUTs.
Similarly, effective extraction of memory-related nodes would greatly benefit FF predictions.
Since LUTs are involved in the entire graph (as computation units and as glue logics to connect the circuit components together), the graph-level understanding is important.
The predictions of SLICE are dependent on FFs and LUTs.
To briefly summarize, it is helpful to carefully characterize neighborhood information from each node's predecessors, successors, itself, and their relations, such that the preference of resource types on different nodes can be clearly understood and the sophisticated mapping rules from heterogeneous nodes to resource usage can be quantitatively learned.
\circled{2} Timing.
Compared with resource predictions, the CP timing predictions show relatively lower error rates and better consistency between DFGs and CDFGs. 
A probable reason is that the CP timing is insensitive to graph sizes since it captures local information between any two FFs, as shown in Fig.~\ref{fig:Program-to-Circuit} (d).
As long as the path segment that introduces the maximum delay is recognized, the CP timing can be accurately predicted.

\subsection{From Synthetic to Real-case Generalization}

\textbf{Generalization Performance.}
Fig.~\ref{fig:realcase} displays the generalization results on real-world applications.
Compared with the test errors on synthetic DFGs and CDFGs, the generalization to real-case shows a large performance degradation.
Several causes to blame on include the divergence in graph sizes, the drastic number of loops, more complicated operations, and appearance of peculiar cases among these real applications.
First, IR graphs in real-world applications generally have a wider range of size, ranging from tens to thousands of nodes/edges. 
This requires GNN models to generalize effectively to much larger graphs than those appearing during the training.
Second, real-world applications involve a considerable number of loops, both in the count of loops and the length of loops.
This easily confuses GNN models to discriminate these graph topologies, thus unable to make proper predictions.
Third, real-world applications come with more complex operations that rarely appear in synthetic programs, which confirms the necessity of our setting with \texttt{misc} in feature embedding and could provoke future investigations on generalization with null values.
Furthermore, these real-world applications can behave wildly and counter-intuitively, where small IR graphs consume much resource yet larger graphs use little resource.
These cases, as well as those extremely large IR graphs, can be recognized as out-of-distribution points.


\textbf{GNN vs. HLS tools.}
GNN predictions are also compared with estimations from commercial HLS tools in Fig.~\ref{fig:realcase}.
Notably, HLS tools estimate resource usage and timing based on RTL designs generated \textit{after} circuit translation (see Fig. \ref{fig:Program-to-Circuit}(a)), taking much longer time;
whereas GNN models make predictions directly from raw IR graphs \textit{before} the translation (see Fig. \ref{fig:Program-to-Circuit}(b)), thus much more challenging.
Fortunately, GNN models can still provide predictions comparable to HLS tools, or generally surpass HLS tools in timing prediction. 
Such results empirically demonstrate that given the current message passing mechanisms and model designs, \textit{GNNs are capable to learn a simplified version of the sophisticated heuristics and mapping rules used in scheduling/binding and logic/physical synthesis, at least equivalent to the rules used in HLS estimations.}

\subsection{Discussion.}
We envision three potential directions that would benefit the Program-to-Circuit problem.

\textbf{Generalization across graph sizes, node degrees, and out-of-distribution cases.}
To impose more impact of GNNs on many EDA problems \cite{khailany2020accelerating}, GNNs are expected to effectively generalize to larger or out-of-distribution graphs, especially in the case that real-world applications greatly vary in sizes.
The Program-to-Circuit problem is a starting task but also a screening test, exhibiting the current difficulty in generalizing across graph sizes and out-of-distribution cases.
We also observe that in IR graphs, while most nodes have low degrees, there do exist several \textit{busy} nodes with high degrees, taking the role of hubs in control and data flows.
It would be interesting for GNNs to further investigate these nodes or scale-free graphs.

\textbf{Structural and algorithmic innovations to process heavy-loop graph topologies.}
There are multiple loops in IR graphs, especially in real-world applications.
From the aspect of domain knowledge, the very large loops usually relate to complex computation operations, which typically involve sharing and interference of computational resources; 
the small loops usually relate to memory operations, which typically involve register allocation and auxiliary computation resources. 
With a large number of small loops, i.e., a large number of memory operations, the register allocation becomes increasingly complicated, since different memory operations could share the same registers in different time steps, or one memory operation may occupy multiple registers simultaneously. The complexity of scheduling and binding of registers grows exponentially, which poses challenges on GNNs to understand these rules. Since memory operations often require auxiliary computations, these operations also have intricate impacts on the allocation of computation resources, which increases the difficulty of accurately estimating resource utilization.
From the aspect of GNN models, the class of message-passing-based GNN models has limited expressiveness and is not better than the 1-Weisfeiler-Lehman isomorphism test \cite{maron2019provably}, which is not good at handling loops.
Given the current limitations, we envisage that\textit{ the discernment of various loops would be beneficial to accurately predict resource usage or other tasks}.

\textbf{ Gap from classification to regression. }
Existing GNN models already show excellent performance on node/graph classification or link prediction tasks, while there is still a long way to go in graph-level regression that demands much higher representation power. 
The Program-to-Circuit problem is a representative requiring predictions of exact values, which will inspire wider applications.

The Program-to-Circuit problem is a pillar stone in the EDA domain.
Given that fast and accurate circuit quality estimation is the foundation for design exploration and optimization in EDA, it is hard to believe success on more challenging tasks would ever happen if we can not properly accomplish the Program-to-Circuit problem.
Thankfully, the current results are not perfect but really promising.

\section{Conclusions}

In this work, we defined and discussed the Program-to-Circuit problem using GNNs, which is of great importance not only in EDA domain but can also largely benefit broader program-related programs, Program-to-X.
We designed a standard benchmark for the Program-to-Circuit problem, experimented 14 different GNNs, and discussed their performance variance with circuit domain-specific analysis. 
We further screened the generality and representation power of GNNs on real-case applications, and discussed possible limitations of current GNN models when applied to program-related problems.
Pioneered in this promising direction, we expect more follow-up benchmarks and studies on GNN representation and generalization capability within the Program-to-X framework.

\bibliographystyle{unsrt}
\bibliography{ref}

\begin{thebibliography}{10}

\bibitem{shlomi2020graph}
Jonathan Shlomi, Peter Battaglia, and Jean-Roch Vlimant.
\newblock Graph neural networks in particle physics.
\newblock {\em Machine Learning: Science and Technology}, 2(2):021001, 2020.

\bibitem{ju2020graph}
Xiangyang Ju, Steven Farrell, Paolo Calafiura, Daniel Murnane, Lindsey Gray,
  Thomas Klijnsma, Kevin Pedro, Giuseppe Cerati, Jim Kowalkowski, Gabriel
  Perdue, et~al.
\newblock Graph neural networks for particle reconstruction in high energy
  physics detectors.
\newblock {\em arXiv preprint arXiv:2003.11603}, 2020.

\bibitem{coley2019graph}
Connor~W Coley, Wengong Jin, Luke Rogers, Timothy~F Jamison, Tommi~S Jaakkola,
  William~H Green, Regina Barzilay, and Klavs~F Jensen.
\newblock A graph-convolutional neural network model for the prediction of
  chemical reactivity.
\newblock {\em Chemical science}, 10(2):370--377, 2019.

\bibitem{gilmer2017neural}
Justin Gilmer, Samuel~S Schoenholz, Patrick~F Riley, Oriol Vinyals, and
  George~E Dahl.
\newblock Neural message passing for quantum chemistry.
\newblock In {\em International Conference on Machine Learning}, pages
  1263--1272. PMLR, 2017.

\bibitem{fan2019graph}
Wenqi Fan, Yao Ma, Qing Li, Yuan He, Eric Zhao, Jiliang Tang, and Dawei Yin.
\newblock Graph neural networks for social recommendation.
\newblock In {\em The World Wide Web Conference}, pages 417--426, 2019.

\bibitem{guo2020deep}
Zhiwei Guo and Heng Wang.
\newblock A deep graph neural network-based mechanism for social
  recommendations.
\newblock {\em IEEE Transactions on Industrial Informatics}, 17(4):2776--2783,
  2020.

\bibitem{zhao2021identifying}
Tianyi Zhao, Yang Hu, Linda~R Valsdottir, Tianyi Zang, and Jiajie Peng.
\newblock Identifying drug--target interactions based on graph convolutional
  network and deep neural network.
\newblock {\em Briefings in bioinformatics}, 22(2):2141--2150, 2021.

\bibitem{lim2019predicting}
Jaechang Lim, Seongok Ryu, Kyubyong Park, Yo~Joong Choe, Jiyeon Ham, and
  Woo~Youn Kim.
\newblock Predicting drug--target interaction using a novel graph neural
  network with 3d structure-embedded graph representation.
\newblock {\em Journal of chemical information and modeling}, 59(9):3981--3988,
  2019.

\bibitem{li2020customized}
Yaguang Li, Yishuang Lin, Meghna Madhusudan, Arvind Sharma, Wenbin Xu, Sachin~S
  Sapatnekar, Ramesh Harjani, and Jiang Hu.
\newblock A customized graph neural network model for guiding analog ic
  placement.
\newblock In {\em 2020 IEEE/ACM International Conference On Computer Aided
  Design (ICCAD)}, pages 1--9. IEEE, 2020.

\bibitem{zhang2019circuit}
Guo Zhang, Hao He, and Dina Katabi.
\newblock Circuit-gnn: Graph neural networks for distributed circuit design.
\newblock In {\em International Conference on Machine Learning}, pages
  7364--7373. PMLR, 2019.

\bibitem{wu2021ironman}
Nan Wu, Yuan Xie, and Cong Hao.
\newblock Ironman: Gnn-assisted design space exploration in high-level
  synthesis via reinforcement learning.
\newblock In {\em Proceedings of the 2021 on Great Lakes Symposium on VLSI},
  pages 39--44, 2021.

\bibitem{khailany2020accelerating}
Brucek Khailany, Haoxing Ren, Steve Dai, Saad Godil, Ben Keller, Robert Kirby,
  Alicia Klinefelter, Rangharajan Venkatesan, Yanqing Zhang, Bryan Catanzaro,
  et~al.
\newblock Accelerating chip design with machine learning.
\newblock {\em IEEE Micro}, 40(6):23--32, 2020.

\bibitem{shi2019learning}
Zhan Shi, Kevin Swersky, Daniel Tarlow, Parthasarathy Ranganathan, and Milad
  Hashemi.
\newblock Learning execution through neural code fusion.
\newblock In {\em International Conference on Learning Representations}, 2019.

\bibitem{ghaffarian2021neural}
Seyed~Mohammad Ghaffarian and Hamid~Reza Shahriari.
\newblock Neural software vulnerability analysis using rich intermediate graph
  representations of programs.
\newblock {\em Information Sciences}, 553:189--207, 2021.

\bibitem{mendis2019ithemal}
Charith Mendis, Alex Renda, Saman Amarasinghe, and Michael Carbin.
\newblock Ithemal: Accurate, portable and fast basic block throughput
  estimation using deep neural networks.
\newblock In {\em International Conference on machine learning}, pages
  4505--4515. PMLR, 2019.

\bibitem{aho2007compilers}
A~Aho, M~Lam, R~Sethi, J~Ullman, Keith Cooper, Linda Torczon, and S~Muchnick.
\newblock Compilers: Principles, techniques and tools.
\newblock 2007.

\bibitem{wolf2012computers}
Marilyn Wolf.
\newblock {\em Computers as components: principles of embedded computing system
  design}.
\newblock Elsevier, 2012.

\bibitem{gcc}
{The GNU Compiler Collection}.
\newblock \url{https://gcc.gnu.org/}.
\newblock [Online; accessed 17-May-2021].

\bibitem{llvm}
{The LLVM Compiler Infrastructure}.
\newblock \url{https://llvm.org/}.
\newblock [Online; accessed 17-May-2021].

\bibitem{CPython}
Cpython.
\newblock \url{https://www.python.org/}.
\newblock [Online; accessed 17-May-2021].

\bibitem{coussy2009introduction}
Philippe Coussy, Daniel~D Gajski, Michael Meredith, and Andres Takach.
\newblock An introduction to high-level synthesis.
\newblock {\em IEEE Design \& Test of Computers}, 26(4):8--17, 2009.

\bibitem{mirhoseini2020chip}
Azalia Mirhoseini, Anna Goldie, Mustafa Yazgan, Joe Jiang, Ebrahim Songhori,
  Shen Wang, Young-Joon Lee, Eric Johnson, Omkar Pathak, Sungmin Bae, et~al.
\newblock Chip placement with deep reinforcement learning.
\newblock {\em arXiv preprint arXiv:2004.10746}, 2020.

\bibitem{wang2020gcn}
Hanrui Wang, Kuan Wang, Jiacheng Yang, Linxiao Shen, Nan Sun, Hae-Seung Lee,
  and Song Han.
\newblock Gcn-rl circuit designer: Transferable transistor sizing with graph
  neural networks and reinforcement learning.
\newblock In {\em 2020 57th ACM/IEEE Design Automation Conference (DAC)}, pages
  1--6. IEEE, 2020.

\bibitem{gao2021layout}
Xiaohan Gao, Chenhui Deng, Mingjie Liu, Zhiru Zhang, David~Z Pan, and Yibo Lin.
\newblock Layout symmetry annotation for analog circuits with graph neural
  networks.
\newblock In {\em Proceedings of the 26th Asia and South Pacific Design
  Automation Conference}, pages 152--157, 2021.

\bibitem{chen2021universal}
Hao Chen, Keren Zhu, Mingjie Liu, Xiyuan Tang, Nan Sun, and David~Z. Pan.
\newblock Universal symmetry constraint extraction for analog and mixed-signal
  circuits with graph neural networks.
\newblock In {\em 2021 58th ACM/IEEE Design Automation Conference (DAC)}, pages
  1--6. IEEE, 2021.

\bibitem{andersen1994program}
Lars~Ole Andersen.
\newblock {\em Program analysis and specialization for the C programming
  language}.
\newblock PhD thesis, Citeseer, 1994.

\bibitem{nielson2004principles}
Flemming Nielson, Hanne~R Nielson, and Chris Hankin.
\newblock {\em Principles of program analysis}.
\newblock Springer Science \& Business Media, 2004.

\bibitem{xue2019machine}
Hongfa Xue, Shaowen Sun, Guru Venkataramani, and Tian Lan.
\newblock Machine learning-based analysis of program binaries: A comprehensive
  study.
\newblock {\em IEEE Access}, 7:65889--65912, 2019.

\bibitem{hennessy2011computer}
John~L Hennessy and David~A Patterson.
\newblock {\em Computer architecture: a quantitative approach}.
\newblock Elsevier, 2011.

\bibitem{barany2017liveness}
Gerg{\"o} Barany.
\newblock Liveness-driven random program generation.
\newblock In {\em International Symposium on Logic-Based Program Synthesis and
  Transformation}, pages 112--127. Springer, 2017.

\bibitem{cuoq2012frama}
Pascal Cuoq, Florent Kirchner, Nikolai Kosmatov, Virgile Prevosto, Julien
  Signoles, and Boris Yakobowski.
\newblock Frama-c.
\newblock In {\em International conference on software engineering and formal
  methods}, pages 233--247. Springer, 2012.

\bibitem{kirchner2015frama}
Florent Kirchner, Nikolai Kosmatov, Virgile Prevosto, Julien Signoles, and
  Boris Yakobowski.
\newblock Frama-c: A software analysis perspective.
\newblock {\em Formal Aspects of Computing}, 27(3):573--609, 2015.

\bibitem{reagen2014machsuite}
Brandon Reagen et~al.
\newblock {MachSuite}: Benchmarks for accelerator design and customized
  architectures.
\newblock In {\em IISWC}, 2014.

\bibitem{hara2009proposal}
Yuko Hara et~al.
\newblock Proposal and quantitative analysis of the chstone benchmark program
  suite for practical c-based high-level synthesis.
\newblock {\em JIP}, 17:242--254, 2009.

\bibitem{PolyBench}
Louis-No{\"e}l Pouchet and Tomofumi Yuki.
\newblock Polybench/c - the polyhedral benchmark suite.
\newblock \url{http://web.cs.ucla.edu/~pouchet/software/polybench/}, 2016.

\bibitem{kipf2016semi}
Thomas~N Kipf and Max Welling.
\newblock Semi-supervised classification with graph convolutional networks.
\newblock {\em International Conference on Learning Representations (ICLR)},
  2016.

\bibitem{wu2019simplifying}
Felix Wu, Amauri Souza, Tianyi Zhang, Christopher Fifty, Tao Yu, and Kilian
  Weinberger.
\newblock Simplifying graph convolutional networks.
\newblock In {\em International conference on machine learning}, pages
  6861--6871. PMLR, 2019.

\bibitem{hamilton2017inductive}
William~L Hamilton, Rex Ying, and Jure Leskovec.
\newblock Inductive representation learning on large graphs.
\newblock In {\em Proceedings of the 31st International Conference on Neural
  Information Processing Systems}, pages 1025--1035, 2017.

\bibitem{bianchi2021graph}
Filippo~Maria Bianchi, Daniele Grattarola, Lorenzo Livi, and Cesare Alippi.
\newblock Graph neural networks with convolutional arma filters.
\newblock {\em IEEE Transactions on Pattern Analysis and Machine Intelligence},
  2021.

\bibitem{ma2020path}
Zheng Ma, Junyu Xuan, Yu~Guang Wang, Ming Li, and Pietro Li{\`o}.
\newblock Path integral based convolution and pooling for graph neural
  networks.
\newblock {\em Advances in Neural Information Processing Systems}, 2020.

\bibitem{xu2018powerful}
Keyulu Xu, Weihua Hu, Jure Leskovec, and Stefanie Jegelka.
\newblock How powerful are graph neural networks?
\newblock {\em International Conference on Learning Representations (ICLR)},
  2019.

\bibitem{corso2020principal}
Gabriele Corso, Luca Cavalleri, Dominique Beaini, Pietro Li{\`o}, and Petar
  Veli{\v{c}}kovi{\'c}.
\newblock Principal neighbourhood aggregation for graph nets.
\newblock {\em Advances in Neural Information Processing Systems}, 2020.

\bibitem{velivckovic2017graph}
Petar Veli{\v{c}}kovi{\'c}, Guillem Cucurull, Arantxa Casanova, Adriana Romero,
  Pietro Lio, and Yoshua Bengio.
\newblock Graph attention networks.
\newblock {\em arXiv preprint arXiv:1710.10903}, 2017.

\bibitem{li2015gated}
Yujia Li, Daniel Tarlow, Marc Brockschmidt, and Richard Zemel.
\newblock Gated graph sequence neural networks.
\newblock {\em International Conference on Learning Representations (ICLR)},
  2016.

\bibitem{schlichtkrull2018modeling}
Michael Schlichtkrull, Thomas~N Kipf, Peter Bloem, Rianne Van Den~Berg, Ivan
  Titov, and Max Welling.
\newblock Modeling relational data with graph convolutional networks.
\newblock In {\em European semantic web conference}, pages 593--607. Springer,
  2018.

\bibitem{gao2019graph}
Hongyang Gao and Shuiwang Ji.
\newblock Graph u-nets.
\newblock In {\em international conference on machine learning}, pages
  2083--2092. PMLR, 2019.

\bibitem{brockschmidt2020gnn}
Marc Brockschmidt.
\newblock Gnn-film: Graph neural networks with feature-wise linear modulation.
\newblock In {\em International Conference on Machine Learning}, pages
  1144--1152. PMLR, 2020.

\bibitem{hu2020ogb}
Weihua Hu, Matthias Fey, Marinka Zitnik, Yuxiao Dong, Hongyu Ren, Bowen Liu,
  Michele Catasta, and Jure Leskovec.
\newblock Open graph benchmark: Datasets for machine learning on graphs.
\newblock {\em arXiv preprint arXiv:2005.00687}, 2020.

\bibitem{Fey/Lenssen/2019}
Matthias Fey and Jan~E. Lenssen.
\newblock Fast graph representation learning with {PyTorch Geometric}.
\newblock In {\em ICLR Workshop on Representation Learning on Graphs and
  Manifolds}, 2019.

\bibitem{xilinx}
Vivado design suite user guide: High-level synthesis (ug902).
\newblock
  \url{https://www.xilinx.com/support/documentation/sw_manuals/xilinx2019_1/ug902-vivado-high-level-synthesis.pdf}.
\newblock [Online; accessed 22-May-2021].

\bibitem{maron2019provably}
Haggai Maron, Heli Ben-Hamu, Hadar Serviansky, and Yaron Lipman.
\newblock Provably powerful graph networks.
\newblock In {\em Proceedings of the 33rd International Conference on Neural
  Information Processing Systems}, pages 2156--2167, 2019.

\end{thebibliography}

\end{document}